\documentclass{llncs}

\usepackage{graphicx}
\graphicspath{{images/}}
\DeclareGraphicsExtensions{.eps}

\usepackage[utf8]{inputenc}

\usepackage[hidelinks]{hyperref}
\usepackage{color}

\usepackage{subcaption}

\usepackage{multirow}
\usepackage[table, dvipsnames]{xcolor}

\usepackage{needspace}
\usepackage{makeidx}  
\usepackage[final]{changes}
\usepackage{todonotes} 

\makeatletter
\setremarkmarkup{\todo[color=Changes@Color#1!10,size=\scriptsize]{#1: #2}}
\setremarkmarkup{\todo[backgroundcolor=Changes@Color#1!20!white,linecolor=Changes@Color#1!30!white,size=\scriptsize\sffamily]{\@nameuse{Changes@AuthorName#1}\\#2}}
\makeatother

\newcommand*{\helvetica}{\fontfamily{phv}\selectfont\scriptsize}

\definechangesauthor[color=Red, name=\textbf{Alejandro}]{AC}

\begin{document}

\frontmatter

\pagestyle{headings}

\title{Recognizing Activities of Daily Living from Egocentric Images}

\author{Alejandro Cartas\inst{1}\thanks{Both authors contributed equally.} \and Juan Marín\inst{1}\textsuperscript{$\star$}\and Petia Radeva\inst{1,2}\and \\Mariella Dimiccoli\inst{1, 2}}

\authorrunning{Cartas et al.}

\institute{University of Barcelona, MAIA Department, 08007 Barcelona, Spain,\\
E-mail: \href{mailto:alejandro.cartas@ub.edu}{alejandro.cartas@ub.edu}, \href{mailto:jmarinve7@alumnes.ub.edu>}{jmarinve7@alumnes.ub.edu}
\and
Computer Vision Center, 08193 Cerdanyola del Vallès, Barcelona, Spain}

\maketitle

\begin{abstract}

Recognizing Activities of Daily Living (ADLs)  has a large number of health applications, such as characterize lifestyle for habit improvement, nursing and rehabilitation services. Wearable cameras can daily gather large amounts of image data that provide rich visual information about ADLs than using other wearable sensors.  In this paper, we explore the classification of ADLs from images captured by low temporal resolution wearable camera (2fpm) by using  a Convolutional Neural Networks (CNN) approach. We show that the classification accuracy of a CNN largely improves
when its output is combined, through a random decision forest, with contextual information from a fully connected layer. The proposed method was tested on a subset of the NTCIR-12 egocentric dataset, consisting of 18,674 images and achieved an overall accuracy of 86\% activity recognition on 21 classes.

\end{abstract}

\section{Introduction}

The Activities of Daily Living (ADLs) include, but are not limited to the activities that an independent person performs on daily basis for living at home or in a community \cite{martin2015design}. The monitoring of these activities on elderly people could prevent health problems \cite{martin2015design,schussler2016potentially}. Recently, egocentric (first-person) cameras have been used to monitor ADLs, because they can provide richer contextual information than using only traditional sensors \cite{Nguyen2016}. These lifelogging devices can generate large volumes of data in matter of days. For instance, a wearable camera such as the Narrative Clip can take more than 2,800 pictures per day from an egocentric (first-person) point of view. To extract information about the behavior of a user, these data needs to be classified in an orderly and timely fashion.

Over the last five years, egocentric activity recognition has been an active area of research. Fathi et al. \cite{fathi2011understanding} presented a probabilistic model that classifies activities from short egocentric videos. Their approach models an activity into a set of different actions, and each action is modeled per frame as a spatio-temporal relationship between the hands and the objects involved on it. They further extend their work \cite{fathi2012learning} by another probabilistic generative model that incorporates the gaze features and that models an action as a sequence of frames. Pirsiavash and Ramanan \cite{pirsiavash2012detecting} introduced a dataset of 18 egocentric actions of daily activities performed by 20 persons in unscripted videos. Furthermore, they presented a temporal pyramid to encode spatio-temporal features along with detected active objects knowledge. These temporal pyramids are the input of support vector machines trained for action recognition. More recently, Ma et al. \cite{Ma_2016_CVPR} proposed a twin stream Convolutional Neural Network (CNN) architecture for activity recognition from videos. One of the streams is used for recognizing the appearance of an object based on a hand segmentation and a region of interest. The other stream recognizes the action using an optical flow sequence. In order to recognize activities, both streams are join and the last layers are finetuned.

\begin{figure}[!t]
\begin{center}
\newcommand\rowspace{0.12cm}
\newcommand\imgscale{0.055}
\newcommand\columnProportion{0.19}
\begin{center}
\begin{minipage}{\columnProportion\columnwidth}
\centering
{\helvetica Biking}
\includegraphics[height=1.5cm]{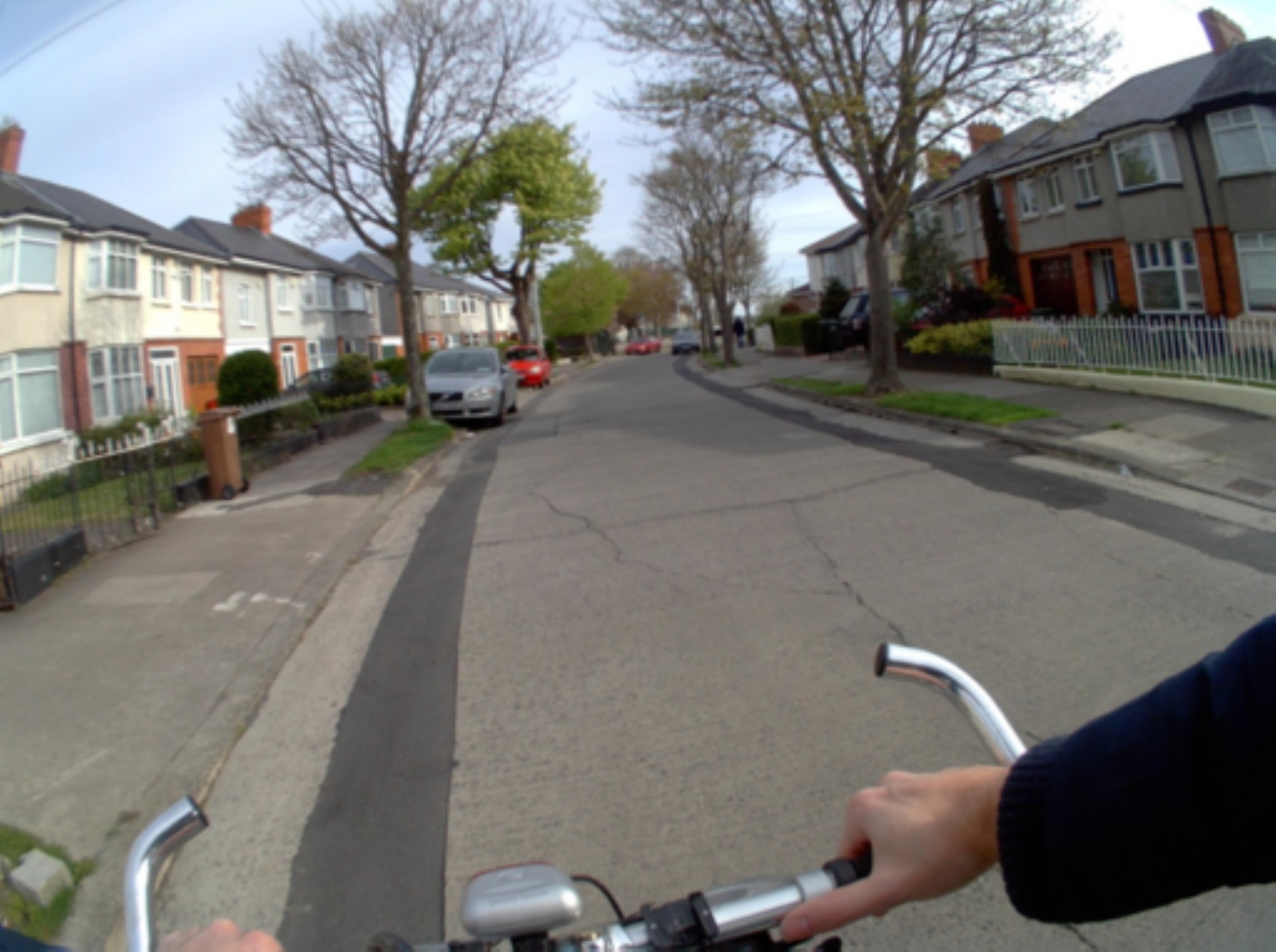}
\end{minipage}%
\begin{minipage}{\columnProportion\columnwidth}
\centering
{\helvetica Driving}
\includegraphics[height=1.5cm]{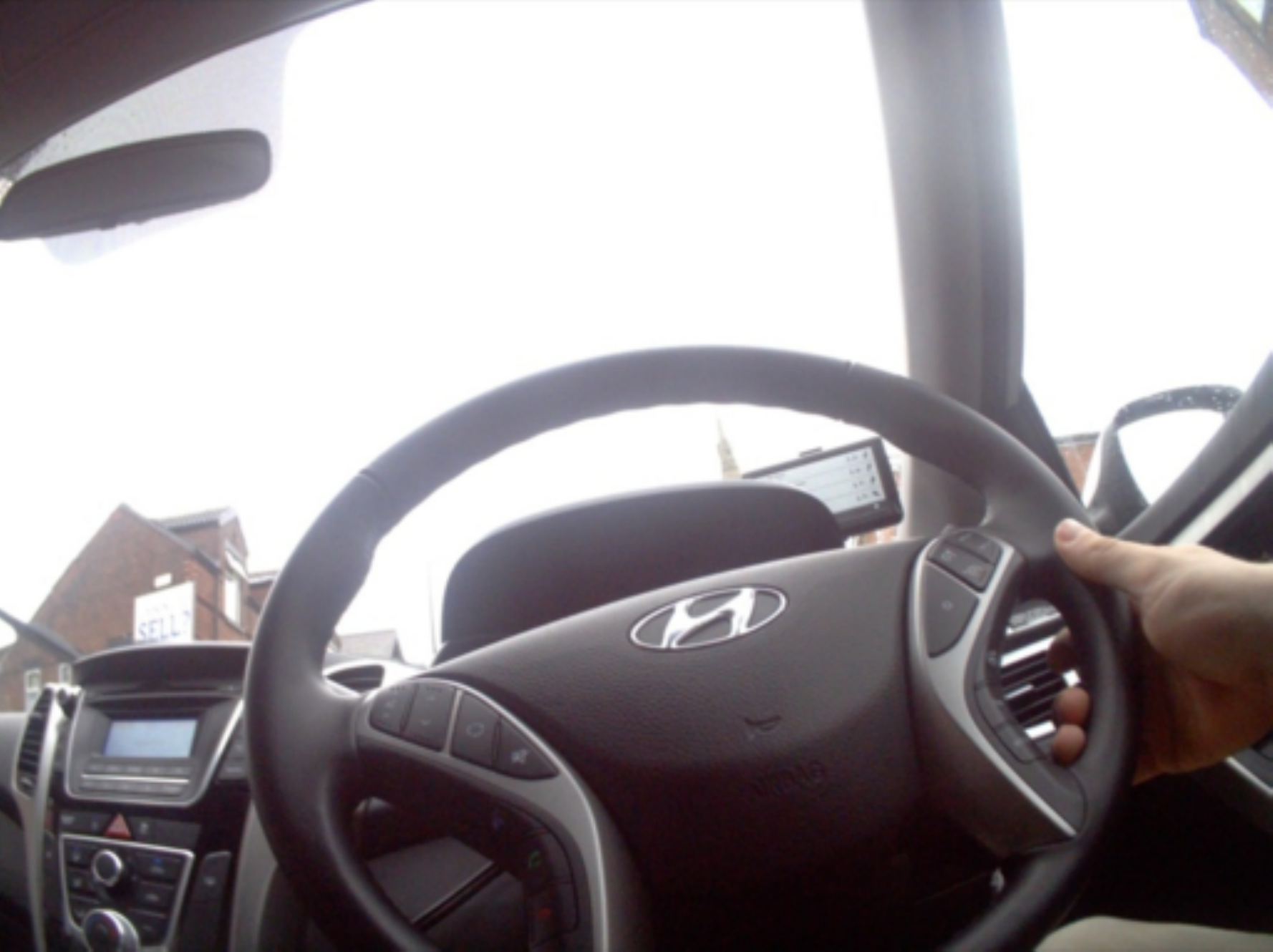}
\end{minipage}%
\begin{minipage}{\columnProportion\columnwidth}
\centering
{\helvetica Talking}
\includegraphics[height=1.5cm]{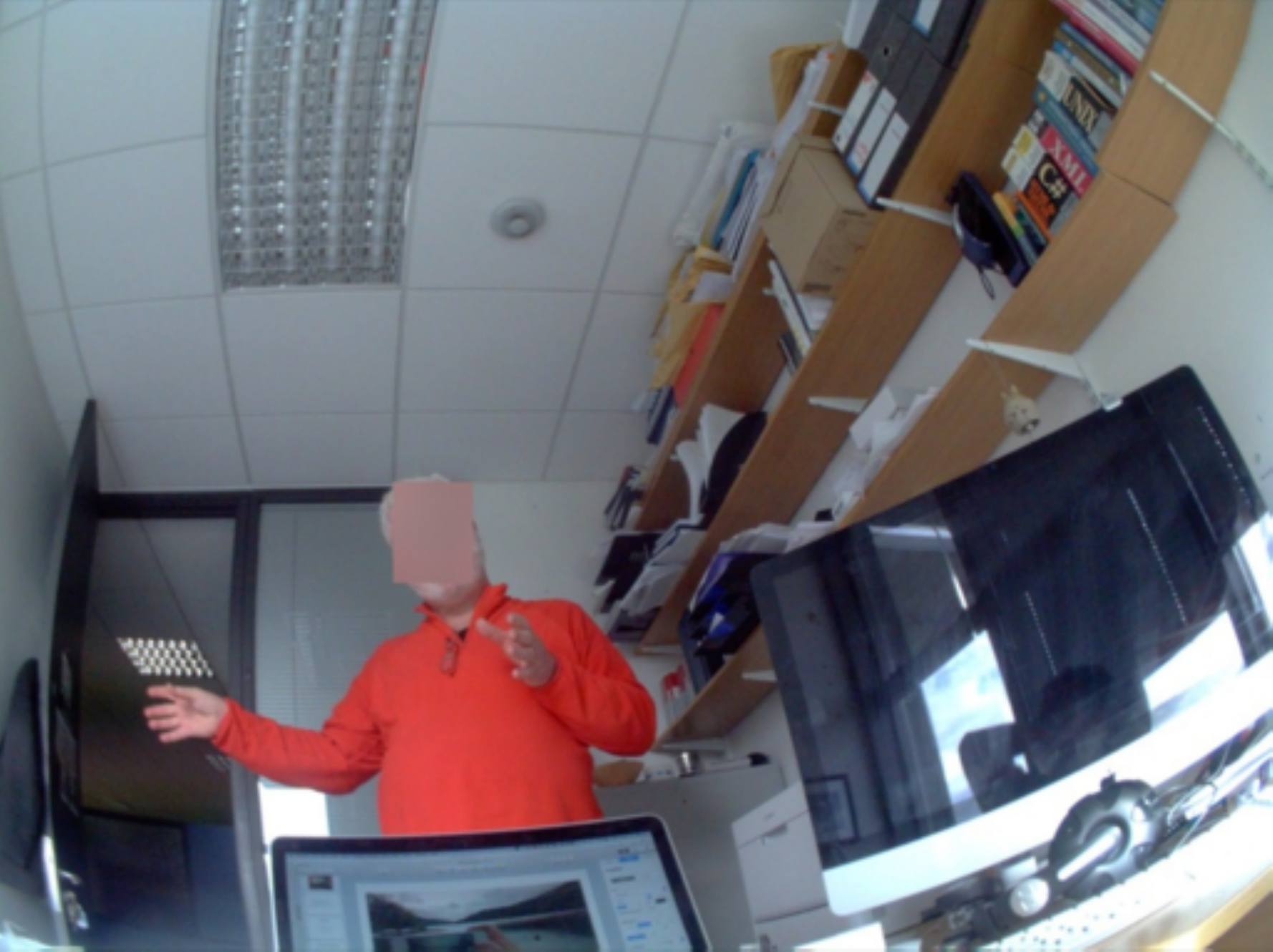}
\end{minipage}%
\begin{minipage}{\columnProportion\columnwidth}
\centering
{\helvetica Plane}
\includegraphics[height=1.5cm]{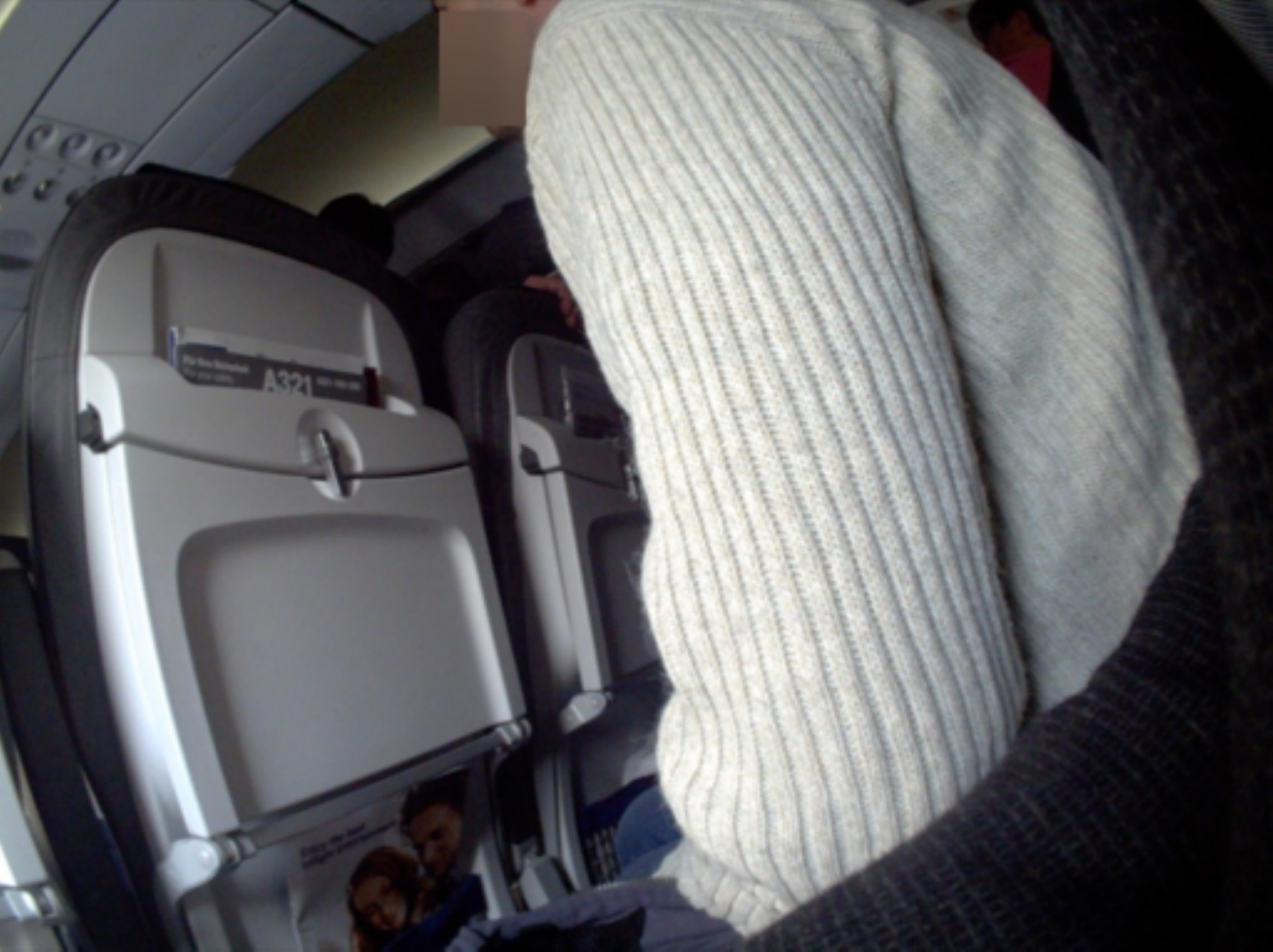}
\end{minipage}%
\begin{minipage}{\columnProportion\columnwidth}
\centering
{\helvetica Shopping}
\includegraphics[height=1.5cm]{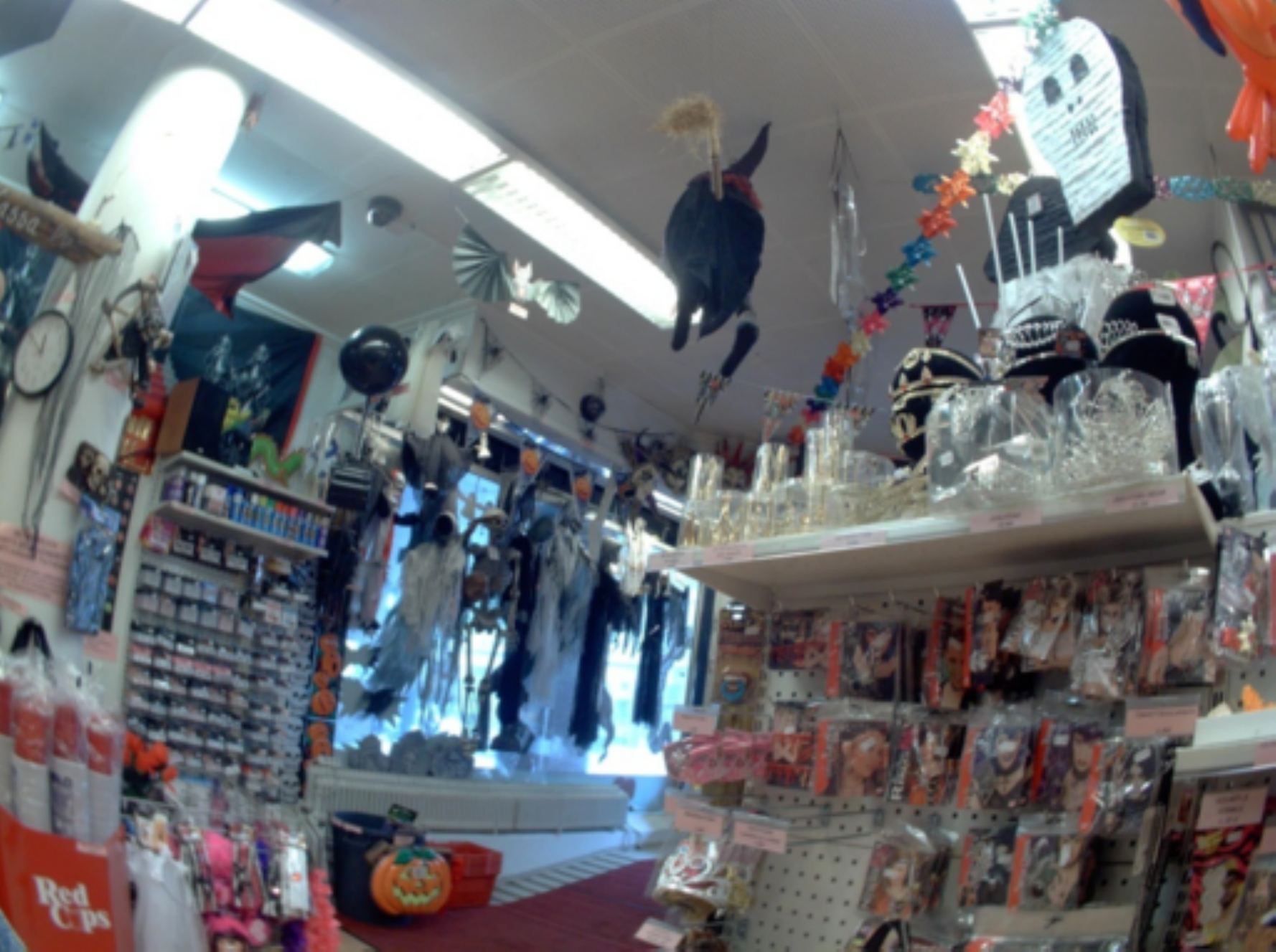}
\end{minipage}\vspace{\rowspace}%

\begin{minipage}{\columnProportion\columnwidth}
\centering
{\helvetica Meeting}
\includegraphics[height=1.5cm]{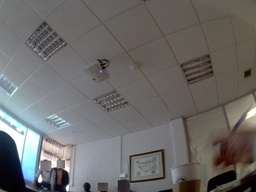}
\end{minipage}%
\begin{minipage}{\columnProportion\columnwidth}
\centering
{\helvetica Cooking}
\includegraphics[height=1.5cm]{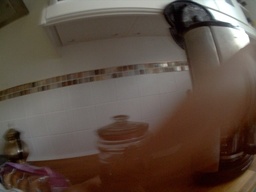}
\end{minipage}%
\begin{minipage}{\columnProportion\columnwidth}
\centering
{\helvetica Cleaning and chores}
\includegraphics[height=1.5cm]{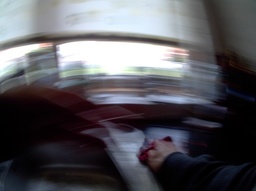}
\end{minipage}%
\begin{minipage}{\columnProportion\columnwidth}
\centering
{\helvetica Drinking together}
\includegraphics[height=1.5cm]{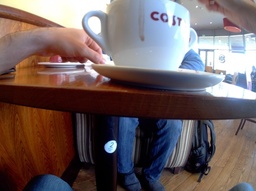}
\end{minipage}%
\begin{minipage}{\columnProportion\columnwidth}
\centering
{\helvetica Eating together}
\includegraphics[height=1.5cm]{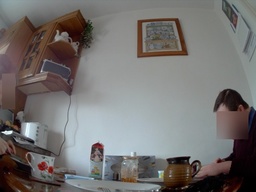}
\end{minipage}\vspace{\rowspace}%

\begin{minipage}{\columnProportion\columnwidth}
\centering
{\helvetica Mobile}
\includegraphics[height=1.5cm]{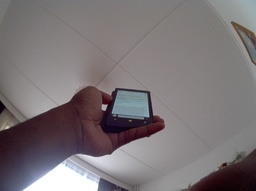}
\end{minipage}%
\begin{minipage}{\columnProportion\columnwidth}
\centering
{\helvetica Public transport}
\includegraphics[height=1.5cm]{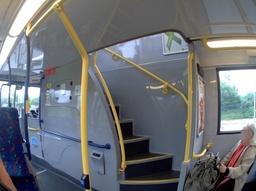}
\end{minipage}%
\begin{minipage}{\columnProportion\columnwidth}
\centering
{\helvetica Reading}
\includegraphics[height=1.5cm]{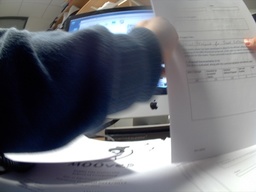}
\end{minipage}%
\begin{minipage}{\columnProportion\columnwidth}
\centering
{\helvetica Working}
\includegraphics[height=1.5cm]{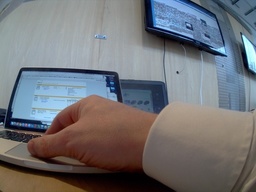}
\end{minipage}%
\begin{minipage}{\columnProportion\columnwidth}
\centering
{\helvetica Socializing}
\includegraphics[height=1.5cm]{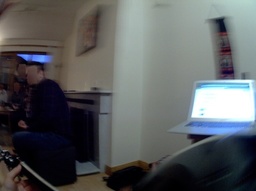}
\end{minipage}%

\end{center}
\caption[]{Examples of annotated images with their corresponding activity label.}
\label{fig:examplesActivities}
\end{center}
\end{figure}

Activity classification from egocentric photo streams is even more difficult problem than from video, since they provide less contextual action information. Castro et. al. \cite{castro2015predicting} introduced a lifelogging dataset composed of 40,103 egocentric images from one subject taken during a 6 months period. The images from this dataset were annotated in 19 different activity categories. In this dataset, the activities performed by the user tend to be routinary and in the same environment. In other words, the activities are almost performed daily at the same time and involved the same objects. Consequently, time and global image features such as color convey useful information for describing activities. To exploit both characteristics, the authors introduced the Late Fusion Ensemble (LTE) method that combines, through a random decision forest, the classification probabilities of a CNN with time and global features, namely color histogram, from the input image. The shortcoming of this approach is that it cannot be generalized to multiple users, since the network need to learn contextual information for each new user. For instance, two distinct persons might have different daily routines, depending on their job, age, etc.

\begin{figure}[!t]
\begin{center}
\includegraphics[scale=0.5]{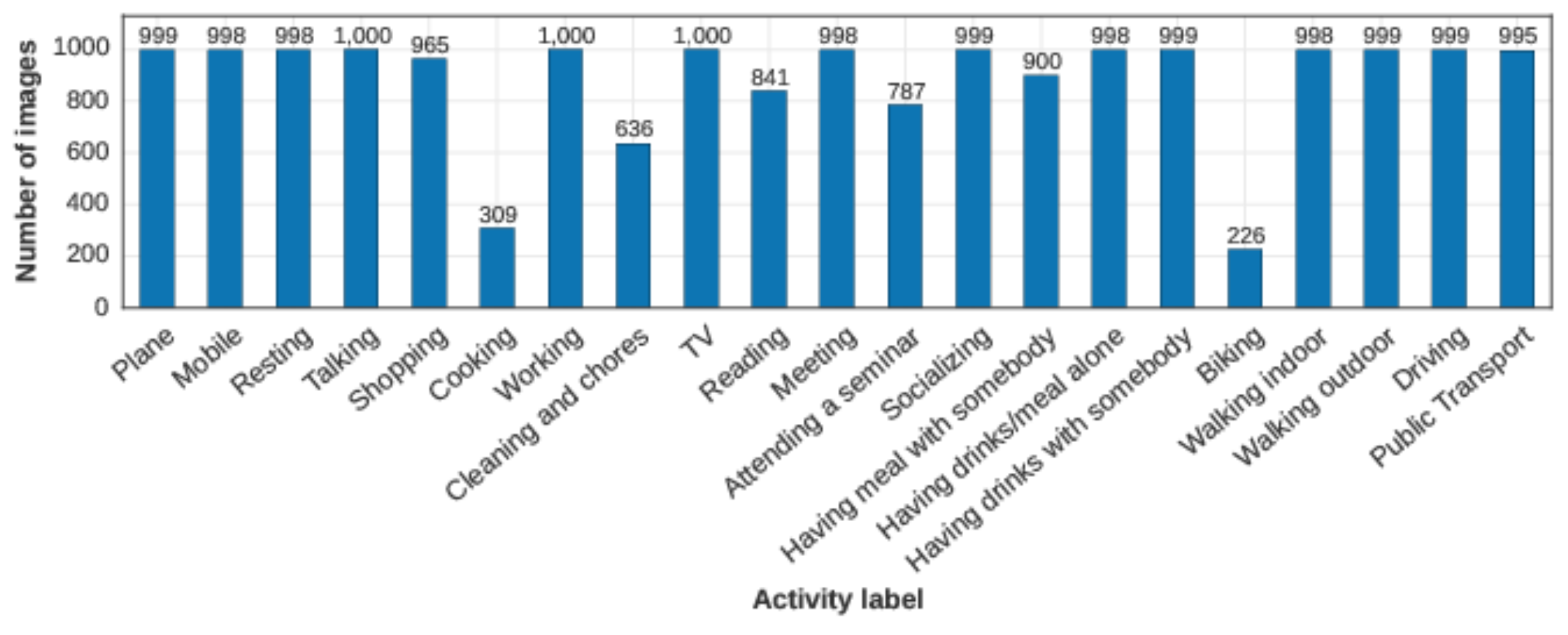}
\caption[]{Our dataset summary.}
\label{fig:overviewNtcir:datasetSummary}
\end{center}
\end{figure}

In this work, we make a step forward on activity recognition from egocentric images by generalizing the task of activity recognition to multiple users. Our approach combines the outputs of different layers of a CNN and use them as the input of a random decision forest. We tested our method on a subset of the NTCIR-12 egocentric dataset \cite{gurrin2016NTCIR}, consisting of 18,674 pictures acquired by three users and annotated with 21 activity labels. Some examples of labeled images with their corresponding category are shown in Fig. \ref{fig:examplesActivities}. Our approach is similar to the LTE introduced in \cite{castro2015predicting}, in the sense that it uses a random forest to combine the output of a CNN with contextual information. However, there are several differences that should be highlighted. First, the problem itself we face is different: instead of classifying the activities of a single user, we classify the activities of threee different users from a dataset having less than half the number of images of the one used in \cite{castro2015predicting}. Hence, our task is more challenging because we must deal with an increased intra-class variability with a much smaller number of images. In addition, since the pictures from the dataset were taken by users with different lifestyles, our method cannot take advantage of their time information and color histogram. To face these problems, instead of using time and color as contextual information, we use the output of a fully connected layer of the CNN.

The rest of the paper is organized as follows: in section \ref{sec:activityClassification} we provide an overview of our approach. In section \ref{sec:dataset}, we introduce the dataset used in the experiments and we briefly describe the annotation tool we created for labeling.  We detail the methodology we followed for conducting our experiments and the different combinations of networks and layers we used in section \ref{sec:methodology}. The results we obtained are discussed in section \ref{sec:results}. Finally, in section \ref{sec:conclusion}, we present our conclusion and final remarks.\vspace{-0.35cm} 

\section{Activity classification}
\label{sec:activityClassification}
\vspace{-0.3cm} 
Our activity classification method is an ensemble composed of a CNN and a random forest. Specifically, our approach joins the output vectors from two layers of a CNN, i.e. the fully connected and the softmax layers, and gives them as input to a random forest. The training of the ensemble is a two-step process. First, a CNN is finetuned and then a random forest is trained over its output vectors. 

The objective of our experiments was to show how the classification performance of the CNN improves by using contextual information. In these experiments, we used two networks as the base of our ensembles, namely the AlexNet \cite{krizhevsky2012AlexNet} and GoogLeNet \cite{szegedy2015GoogLeNet}. Both neural networks were finetuned on a subset of images from the NTCIR-12 dataset \cite{gurrin2016NTCIR} annotated for activity classification.

\section{Dataset}
\label{sec:dataset}

In our experiments, we used a subset of images from the NTCIR-12 dataset \cite{gurrin2016NTCIR} that consists of 89,593  egocentric pictures belonging to three persons. Each user worn a first-person camera in a period of almost a month, totaling 79 days. This camera passively took two pictures per minute.

The dataset we used in this work, consists of a subset of 18,674 images from the NTCIR-12 dataset. These pictures correspond to all users and have different dates and times. We used 21 activity categories to label them using our annotation tool, which is briefly detailed in the next subsection. Although our subset of annotated images is imbalanced, less than the half of categories have less instances than the rest. Some examples of annotated activities and the distribution of the number of images by category are shown in Fig. \ref{fig:examplesActivities} and Fig. \ref{fig:overviewNtcir:datasetSummary}, respectively. The dataset was split in 13,991 training images, 1,857 validation images, and 2,796 testing images. This split maintained the same proportion of examples per category and was used in all our experiments.

\subsubsection{Annotation tool}
\label{sec:annotationTool}

We created a web-based annotation tool specifically target for labeling large amounts of images. The following design guidelines were considered:

\begin{itemize}
\item \textit{Easy interaction}. Browsing and annotating large number of images can be done in an intuitive way. Additionally, descriptive tags can be individually added to a specific picture.
\item \textit{Speed and performance}. The tool can handle several connected users at the same time and present their large collection of pictures.
\item \textit{Privacy and security}. Personal pictures of a user are maintained private. 
\end{itemize}

\begin{table}[!t]
\centering
\resizebox{1.0\columnwidth}{!}{%
\begin{tabular}{ | l | c |c |c |c |c |c |c |}
\hline
\multicolumn{1}{|c|}{\multirow{2}{*}{\textbf{Activity}}} & \multirow{2}{*}{\textbf{AlexNet}} & \textbf{AlexNet+RF} & \textbf{AlexNet+RF} & \textbf{AlexNet+RF} & \multirow{2}{*}{\textbf{GoogLeNet}} & \textbf{GoogLeNet+RF} & \textbf{GoogLeNet+RF} \\
 &  & \textbf{on FC6} & \textbf{on Prob}  & \textbf{on FC6+Prob} &  & \textbf{on Prob} & \textbf{on pool5/7x7+prob} \\ \hline
Public Transport & 82.17 &  \textbf{87.60}  & 86.05 & 86.82 & 79.07 & 80.62 & 84.50 \\ \hline
Driving &  \textbf{100.00}  &  \textbf{100.00}  &  \textbf{100.00}  &  \textbf{100.00}  &  \textbf{100.00}  &  \textbf{100.00}  &  \textbf{100.00}  \\ \hline
Walking outdoor & 84.52 & 86.31 &  \textbf{88.10}  & 86.31 & 83.93 & 85.12 &  \textbf{88.10}  \\ \hline
Walking indoor & 61.88 & 68.75 & 65.00 &  \textbf{71.88}  & 60.00 & 63.75 & 68.75 \\ \hline
Biking &  \textbf{81.58}  & 73.68 & 78.95 & 73.68 & 68.42 & 71.05 & 76.32 \\ \hline
Drinking together & 84.09 &  \textbf{90.91}  & 82.58 & 87.88 & 76.52 & 80.30 & 87.88 \\ \hline
Drinking/eating alone & 70.89 &  \textbf{84.81}  & 74.05 & 80.38 & 60.13 & 70.25 & 75.95 \\ \hline
Eating together & 76.56 &  \textbf{86.72}  & 80.47 & 82.81 & 75.78 & 82.03 & 85.16 \\ \hline
Socializing & 65.31 &  \textbf{89.80}  & 77.55 & 83.67 & 63.95 & 72.11 & 79.59 \\ \hline
Attending a seminar & 89.92 &  \textbf{93.28}  & 89.08 &  \textbf{93.28}  & 82.35 & 86.55 & 89.08 \\ \hline
Meeting & 78.01 & 85.82 & 80.14 & 85.11 & 73.76 & 82.98 &  \textbf{87.94}  \\ \hline
Reading & 69.40 & 84.33 & 77.61 & 82.09 & 85.82 & 83.58 &  \textbf{86.57}  \\ \hline
TV & 88.74 & 95.36 & 92.72 & 94.04 & 94.70 & 96.03 &  \textbf{98.01}  \\ \hline
Cleaning and chores & 44.74 & 61.40 & 56.14 & 71.93 & 51.75 & 67.54 &  \textbf{77.19}  \\ \hline
Working & 89.24 &  \textbf{95.57}  & 91.14 &  \textbf{95.57}  & 90.51 & 91.77 & 94.94 \\ \hline
Cooking & 31.11 & 46.67 &  \textbf{48.89}  & 42.22 & 24.44 & 37.78 & 44.44 \\ \hline
Shopping & 73.88 &  \textbf{85.07}  & 76.12 & 81.34 & 76.87 & 76.87 & 78.36 \\ \hline
Talking & 70.47 & 82.55 & 77.85 & 81.88 & 71.81 & 78.52 &  \textbf{83.89}  \\ \hline
Resting & 98.70 &  \textbf{99.35}  & 98.70 &  \textbf{99.35}  & 97.40 & 98.70 &  \textbf{99.35}  \\ \hline
Mobile & 79.19 & 85.91 & 85.23 &  \textbf{88.59}  & 79.87 & 84.56 & 86.58 \\ \hline
Plane & 90.91 & 93.01 & 91.61 & 90.91 & 93.71 & 94.41 &  \textbf{96.50}  \\ \hline
\noalign{\vskip 0.25cm} \hline
\textbf{Accuracy} &  78.51   &  \textbf{86.58}   &  82.36   &  85.79   &  78.11   &  82.22   &  86.04   \\ \hline
\textbf{Macro precision} &  78.46   &  \textbf{87.29}   &  81.57   &  86.83   &  77.17   &  81.55   &  86.20   \\ \hline
\textbf{Macro recall} &  76.73   &  \textbf{84.61}   &  80.86   &  83.80   &  75.75   &  80.22   &  84.24   \\ \hline
\textbf{Macro F1-score} &  77.09   &  \textbf{85.45}   &  81.11   &  84.68   &  76.06   &  80.64   &  84.84   \\ \hline 
\end{tabular}
}
\caption{Comparison of the ensembles of CNN+Random forest on different combinations of layers. Upper table shows the recall per class and the lower table shows the performance metrics.}
\label{tab:classificationComparison}
\end{table}

\section{Methodology}
\label{sec:methodology}

The main objective of the experiments was to determine if contextual information can improve the activity classification accuracy of a CNN by ensembling the CNN with a random forest. Additionally, we tried to find out the ensemble with the best combination of layers for performance improvement. For training and testing purposes we used the dataset split described in the last section. Since the dataset was imbalanced, we assessed the performance of the ensembles by not only using the accuracy, but also macro metrics for precision, recall, and F1-score. The next subsection presents all the proposed ensemble configurations with their respective training procedure.

\subsection{Ensemble configurations}

In order to train our ensembles, we first fine-tuned AlexNet and GoogLeNet on our dataset. Both models were fine-tuned using the Caffe framework \cite{jia2014caffe} with the same number of iterations (approx. 10 epochs) and the following settings:

\begin{itemize}
\item \textbf{AlexNet}. It was trained using stochastic gradient descent for 2,180 iterations with a batch size of 64 images, a learning rate $\alpha=3$x$10^{-5}$, and a momentum $\mu=0.9$.
\item \textbf{GoogLeNet}. It was trained for 4,370 iterations with a batch size of 32 images, a learning rate $\alpha=3.7$x$10^{-5}$, and a momentum $\mu=0.9$.
\end{itemize}

After fine-tuning our baseline networks, we train several ensembles of random forests combining different final output layers of the CNNs. The following combinations were used in our experiments:

\begin{itemize}
\item \textbf{AlexNet+RF on Prob}. This configuration was trained using a random forest of 500 trees on the softmax probability layer.
\item \textbf{GoogLeNet+RF on Prob}. This configuration was trained using a random forest of 500 trees on the softmax probability layer.
\item \textbf{AlexNet+RF on FC6}. We trained a random forest of 500 trees on only the first fully connected layer (FC6), containing a vector of 4,096 elements.
\item \textbf{AlexNet+RF on FC6+Prob}. A random forest of 500 trees was trained on the FC6 and the softmax layers of AlexNet, thus summing a vector of 4,117 elements.
\item \textbf{GoogLeNet+RF on Pool5/7x7+Prob}. This configuration was trained using a random forest of 500 trees on the pool5/7x7 fully connected layer and the softmax probability layer, thus having a vector of 1,045 features.
\end{itemize}

\begin{figure}[!t]
\begin{center}
\includegraphics[scale=0.44]{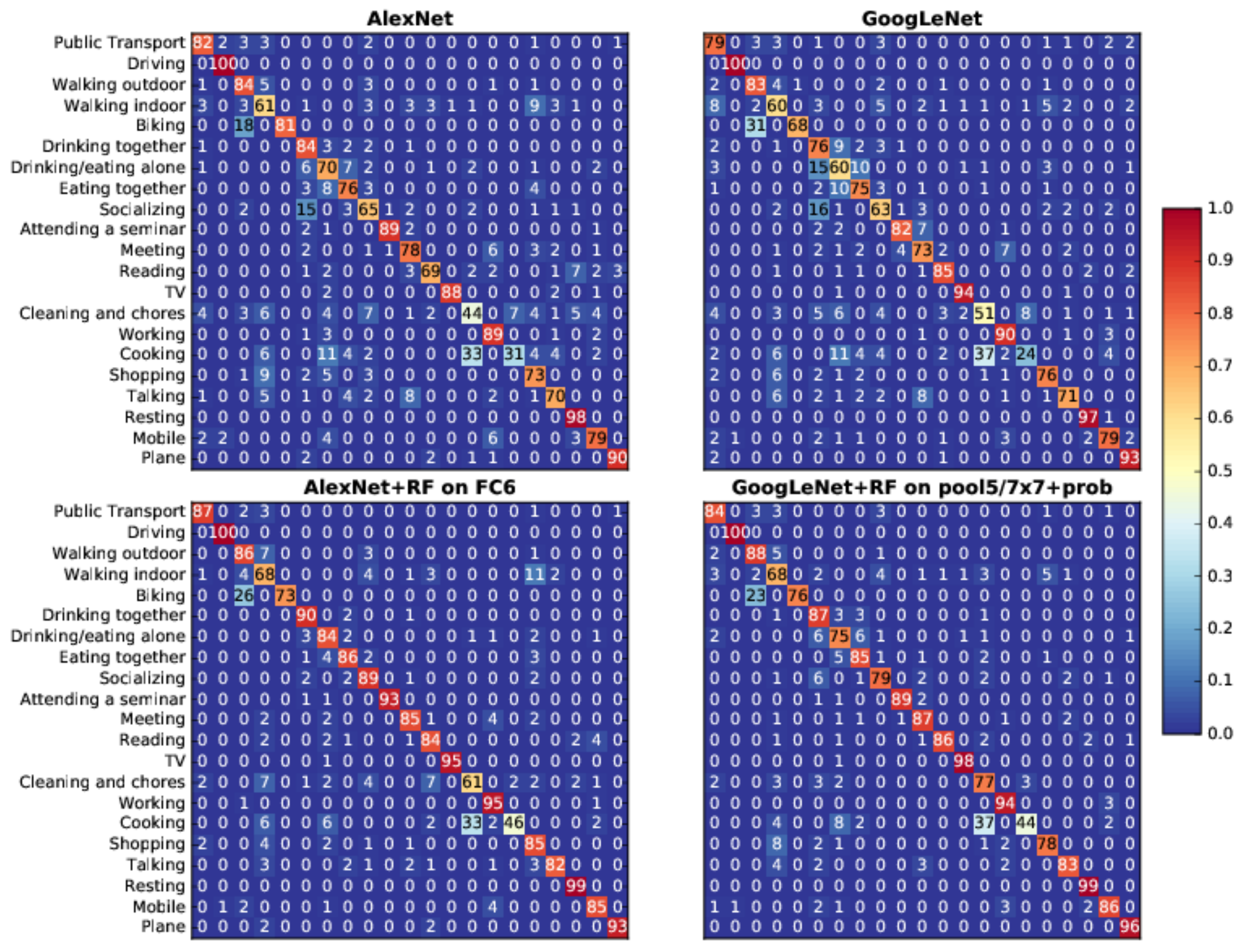}
\caption[]{Normalized confusion matrices of the best combination of layers for each baseline convolutional neural network. This figure is best seen in color.}
\label{fig:confusion_matrices}
\end{center}
\end{figure}

\begin{figure}[!t]
\newcommand\imgscale{0.08}
\newcommand\spaceBetweenBoxes{0.3cm}
\newcommand\columnProportion{0.9}
\centering
\begin{minipage}{.3\textwidth}
\centering
\vspace{-0.7cm}
\begin{minipage}{\columnProportion\textwidth}
\centering
{\helvetica Drinking/eating alone}
\includegraphics[height=2cm]{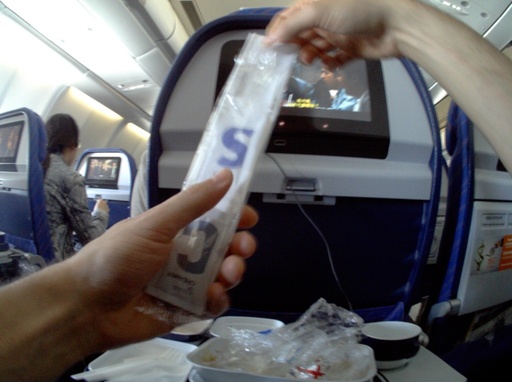}
\end{minipage}\vspace{\spaceBetweenBoxes}

\begin{minipage}{\columnProportion\textwidth}
\centering
{\helvetica AlexNet Top 5}
\resizebox{1.0\columnwidth}{!}{%
\begin{tabular}{|c|c|c|}
\hline
\textbf{\#} & \textbf{Activity} & \textbf{Score} \\ \hline
1 & Public Transport & 0.1864 \\ \hline
2 & Cooking & 0.1464\\ \hline
3 & Eating together & 0.1382\\ \hline
\cellcolor{green!25} \textbf{4} & \cellcolor{green!25} \textbf{Drinking/eating alone} & \cellcolor{green!25} \textbf{0.1223}\\ \hline
5 & Cleaning and chores & 0.1067\\ \hline
\end{tabular}
}
\end{minipage}\vspace{\spaceBetweenBoxes}
    
\begin{minipage}{\columnProportion\textwidth}
\centering
{\helvetica GoogLeNet Top 5}
\resizebox{1.0\columnwidth}{!}{%
\begin{tabular}{|c|c|c|}
\hline
\textbf{\#} & \textbf{Activity}  & \textbf{Score} \\ \hline
1 & Plane & 0.2004\\ \hline
2 & Public Transport & 0.1943\\ \hline
3 & Cleaning and chores & 0.1450\\ \hline
4 & Cooking & 0.0925\\ \hline
\cellcolor{green!25} \textbf{5} & \cellcolor{green!25} \textbf{Drinking/eating alone} & \cellcolor{green!25} \textbf{0.0748}\\ \hline
\end{tabular}
}
\end{minipage}
\end{minipage}%
\begin{minipage}{.3\textwidth}
\centering
\vspace{-0.2cm}
\begin{minipage}{\columnProportion\textwidth}
\centering
{\helvetica Meeting}
\includegraphics[height=2cm]{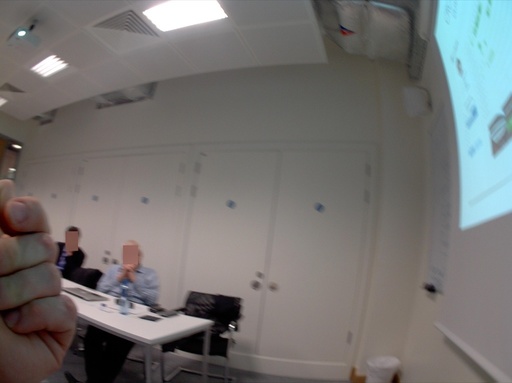}
\end{minipage}\vspace{\spaceBetweenBoxes}
    
\begin{minipage}{\columnProportion\textwidth}
\centering
{\helvetica AlexNet Top 5}
\resizebox{1.0\textwidth}{!}{%
\begin{tabular}{|c|c|c|}
\hline
\textbf{\#} & \textbf{Activity}  & \textbf{Score} \\ \hline    
1 & Talking & 0.2034 \\ \hline
\cellcolor{green!25} \textbf{2} & \cellcolor{green!25} \textbf{Meeting} & \cellcolor{green!25} \textbf{0.1701} \\ \hline
3 & Cooking & 0.1090 \\ \hline
4 & Cleaning and chores & 0.0903 \\ \hline
5 & Shopping & 0.0878 \\ \hline
\end{tabular}
}
\end{minipage}\vspace{\spaceBetweenBoxes}
    
\begin{minipage}{\columnProportion\textwidth}
\centering
{\helvetica GoogLeNet Top 5}
\resizebox{1.0\textwidth}{!}{%
\begin{tabular}{|c|c|c|}
\hline
\textbf{\#} & \textbf{Activity}  & \textbf{Score} \\ \hline
\cellcolor{red!25} \textbf{1} & \cellcolor{red!25} \textbf{Eating together} & \cellcolor{red!25} \textbf{0.3065} \\ \hline
2 & Talking & 0.1215  \\ \hline
3 & Socializing & 0.1062 \\ \hline
4 & Cleaning and chores & 0.08126 \\ \hline
5 & Meeting & 0.0664  \\ \hline
\end{tabular}
}
\end{minipage}
\end{minipage}%
\begin{minipage}{.3\textwidth}
\centering
\vspace{-0.9cm}
\begin{minipage}{\columnProportion\textwidth}
\centering
{\helvetica Reading}
\includegraphics[height=2cm]{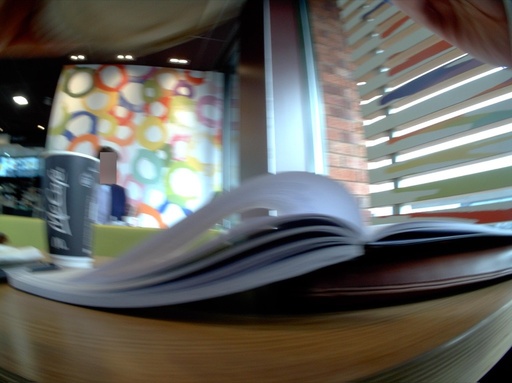}
\end{minipage}\vspace{\spaceBetweenBoxes}
    
\begin{minipage}{\columnProportion\textwidth}
\centering
{\helvetica AlexNet Top 5}
\resizebox{1.0\textwidth}{!}{%
\begin{tabular}{|c|c|c|}
\hline
\textbf{\#} & \textbf{Activity}  & \textbf{Score} \\ \hline    

\cellcolor{red!25} \textbf{1} & \cellcolor{red!25} \textbf{Drinking/eating alone} & \cellcolor{red!25} \textbf{0.5002} \\ \hline
2 & Cleaning and chores & 0.1511748880 \\ \hline
3 & Eating together & 0.1263086796 \\ \hline
4 & Shopping & 0.0589886233 \\ \hline
5 & Drinking together & 0.0251834411 \\ \hline
\end{tabular}
}
\end{minipage}\vspace{\spaceBetweenBoxes}
    
\begin{minipage}{\columnProportion\textwidth}
\centering
{\helvetica GoogLeNet Top 5}
\resizebox{1.0\textwidth}{!}{%
\begin{tabular}{|c|c|c|}
\hline
\textbf{\#} & \textbf{Activity}  & \textbf{Score} \\ \hline
\cellcolor{red!25} \textbf{1} & \cellcolor{red!25} \textbf{Cleaning and chores} & \cellcolor{red!25} \textbf{0.4259} \\ \hline
2 & Eating together & 0.1145323068 \\ \hline
3 & Drinking/eating alone & 0.1137253270 \\ \hline
4 & Drinking together & 0.0841688886 \\ \hline
5 & Reading & 0.06203 \\ \hline
\end{tabular}
}	
\end{minipage}
\end{minipage}
\caption[]{Classification activity examples. On top of each image is shown its true activity label and on bottom its top 5 predictions by AlexNet and GoogLeNet. Additionally, the result of the ensembles \textit{AlexNet+RF on FC6\deleted[id=AC]{+prob}} and \textit{GoogLeNet+RF on Pool5/7x7+prob} is highlighted on color in its corresponding table. The green and red colors means true positive and false positive classification, respectively.} 
\label{fig:classification_examples}\vspace{0.5cm}
\end{figure}

\section{Results}
\label{sec:results}

Our experiments show that the ensemble of CNN plus a random forest improves the performance for both baseline CNN. Table \ref{tab:classificationComparison} shows the classification performance on the the baseline CNN and on different ensembles. The baseline CNN have a similar performance. After using the random forest on the output the softmax layer, the accuracy improved around \replaced[id=AC]{4\%}{10\%}. Specifically, the recall of \added[id=AC]{some} categories with fewer learning instances improved significantly, such as \textit{Cooking}, and \textit{Cleaning and Choring}. Additionally, high overlapping classes such as \textit{Drinking/Eating alone} and \textit{Eating together} also improved their accuracy. The improvement over the overlapping classes can also be seen the confusion matrices shown in Fig. \ref{fig:confusion_matrices}. This means that the random forest improved the classification of images belonging to categories that score similar probabilities. \added[id=AC]{Moreover, the only decrease on accuracy is presented on the class \textit{Biking}}. Since its accuracy on the baseline CNN is very high ($81.68\%$) considering the small number of learning instances (226), we believe this decrease is \replaced[id=AC]{a consequence of}{ due to the fact that} the random forest trying to balance the prediction error among classes.

{\fussy The results of adding contextual information from fully connected layers show a better performance on classification. Table \ref{tab:classificationComparison} shows that the best ensembles were the \textit{AlexNet+RF on FC6\deleted[id=AC]{+prob}} and \textit{GoogLeNet+RF on Pool5/7x7+prob}. Furthermore, these ensembles improved the baseline accuracy by \replaced[id=AC]{8.07\%}{17.85\%} and \replaced[id=AC]{7.93\%}{17.49\%} for AlexNet and GoogLeNet, respectively. Although the performance metrics improved decreasingly with respect to the ensembles that only used the softmax layer, the extra features removed the overfitting problem on the \textit{Walking outdoor} category. Some classification examples are shown in Fig. \ref{fig:classification_examples}.}\looseness=-1

\section{Conclusion}
\label{sec:conclusion}

We presented an egocentric activity classifier ensemble method, which combines different layers of a CNN through a random forest. Specifically, the random forest takes as input a vector containing the output of the softmax probability layer and a fully connected layer encoding global image features. We tested several ensembles based on AlexNet and GoogLeNet, achieving a \replaced[id=AC]{8\%}{17\%} performance improvement on our dataset. The proposed method has been tested on a subset of the NTCIR-12 egocentric dataset consisting of 18,674 images that we labeled with 21 different activity labels. Although we obtained $86\%$ accuracy on a quite difficult task, we believe that there is still room for improvement. Indeed, the proposed approach operates at image-level, without taking into account the temporal coherence of photo-streams. Future work will investigate how to take temporal coherence into account. 

{\footnotesize
\section*{Acknowledgments} 

A.C. was supported by a doctoral fellowship from the Mexican Council of Science and Technology (CONACYT) (grant-no. 366596). This work was partially founded by TIN2015-66951-C2, SGR 1219, CERCA, \textit{ICREA Academia'2014} and 20141510 (Marat\'{o} TV3). The funders had no role in the study design, data collection, analysis, and preparation of the manuscript. M.D. is grateful to the NVIDIA donation program for its support with GPU card.}

\bibliographystyle{abbrv}
\bibliography{activity} 
\end{document}